\documentclass[final]{l4dc2024}

\title[Expert with Clustering: Hierarchical Online Preference Learning Framework]{Expert with Clustering: Hierarchical Online Preference Learning Framework}

\usepackage{times}
\usepackage[utf8]{inputenc} 
\usepackage[T1]{fontenc}    
\usepackage{hyperref}       
\usepackage{url}            
\usepackage{booktabs}       
\usepackage{amsfonts}       
\usepackage{nicefrac}       
\usepackage{microtype}      
\usepackage{xcolor}         
\usepackage{algorithm}
\usepackage{algorithmic}
\usepackage{float}
\usepackage{graphicx}
\usepackage{caption}
\usepackage{subcaption}
\usepackage{placeins}
\usepackage[english]{babel}
\usepackage[utf8]{inputenc}
\usepackage{mathrsfs}     
\usepackage{bbm}

\DeclareMathOperator*{\argmin}{\arg\!\min}

\newcommand{\jhedit}[1]{\textcolor{black}{#1}}

\newcommand{\tyedit}[1]{\textcolor{black}{#1}}

\newcommand{\revise}[1]{\textcolor{black}{#1}}

\author{%
 \Name{Tianyue Zhou} \Email{zhouty1@shanghaitech.edu.cn}\\
 \addr ShanghaiTech University
 \AND
 \Name{Jung-Hoon Cho} \Email{jhooncho@mit.edu}\\
 \addr Massachusetts Institute of Technology
 \AND
 \jhedit{
 \Name{Babak Rahimi Ardabili} \Email{brahimia@charlotte.edu}\\
 \Name{Hamed Tabkhi} \Email{htabkhiv@charlotte.edu}\\
 \addr University of North Carolina at Charlotte
 }
 \AND
 \Name{Cathy Wu} \Email{cathywu@mit.edu}\\
 \addr Massachusetts Institute of Technology
}

\begin{document}
\maketitle
\vspace{-0.15in}
\begin{abstract}
Emerging mobility systems are increasingly capable of recommending options to users, to guide them towards personalized yet sustainable system outcomes. Even more so than the typical recommendation system, it is crucial to minimize regret, because 1) the mobility options directly affect the lives of the users, and 2) the system sustainability relies on sufficient user participation.
In this study, we thus consider accelerating user preference learning by exploiting a low-dimensional space that captures the mobility preferences of users within a population. We therefore 
introduce a hierarchical contextual bandit framework named Expert with Clustering (EWC), which integrates clustering techniques and prediction with expert advice. EWC efficiently utilizes hierarchical user information and incorporates a Loss-guided Distance metric. This metric is instrumental in generating more representative cluster centroids, thereby enhancing the performance of recommendation systems. 
In a recommendation scenario with \(N\) users, \(T\) rounds per user, and \(K\) options, our algorithm achieves a regret bound of \(O(N\sqrt{T\log K} + NT)\). This bound consists of two parts:
the regret from the Hedge algorithm, and 
the average loss from clustering. 
\revise{To the best of the authors knowledge, this is the first work to analyze the regret of an integrated expert algorithm with k-Means clustering.}
This regret bound underscores the theoretical and experimental efficacy of EWC.
Experimental results highlight that EWC can substantially reduce regret by 27.57\% compared to the LinUCB baseline. 
Our work offers a data-efficient approach to capturing both individual and collective behaviors, making it highly applicable to contexts with latent hierarchical structures. 
We expect the algorithm to be applicable to other settings with layered nuances of user preferences and information.
\end{abstract}
\vspace{-0.1in}
\section{Introduction}
Emerging mobility systems are increasingly pivotal in designing efficient and sustainable transit networks. 
These systems with advanced technologies have the potential to revolutionize how we navigate urban environments, thereby enhancing the overall efficiency of transportation systems. 
Recent work aims to minimize emissions through eco-driving recommendation systems, thus contributing to environmental sustainability (\cite{tu_effective_2022, chada_evaluation_2023}).

The challenge lies in the complex nature of drivers' preferences, which are shaped by various factors such as personal schedules, environmental concerns, and the unpredictability of human behavior. 
To tackle this, we advocate for the contextual bandit algorithm, a framework adept at learning from and adapting to the driver's unique context. This algorithm is poised to accurately capture the subtle preferences of drivers, offering nuanced insights into their decision-making processes in ever-changing environments. 

Our problem deviates from a classical contextual bandit scenario. It resembles supervised online learning, as the driver’s choice is known after the system provides recommendations. However, this choice may be influenced by the recommendation itself, leading to varying observed rewards. This variation closely mirrors the structure of a bandit problem, where actions influence observed outcomes. Thus, our problem can be considered a specialized variant of the bandit problem, incorporating elements of both supervised learning and adaptive decision-making typical of bandit scenarios.
Our research focuses on refining online learning algorithms tailored to the contextual bandit framework, enhancing their ability to discern traveler preferences and predict responses to routing and transit mode recommendations. 

\tyedit{In this paper, we propose the Expert with Clustering (EWC) framework, a novel approach that synergizes clustering and prediction with expert advice. The fundamental concept involves using clustering to discern hierarchical information among users, with each cluster acting as an `expert' representing common user preferences. This approach transforms the online preference learning problem into one of prediction with expert advice. For each user, an expert is selected to approximate their preferences, enabling accurate recommendations.}
\begin{figure}[!t]
    \centering    
    \includegraphics[width=0.9\textwidth]{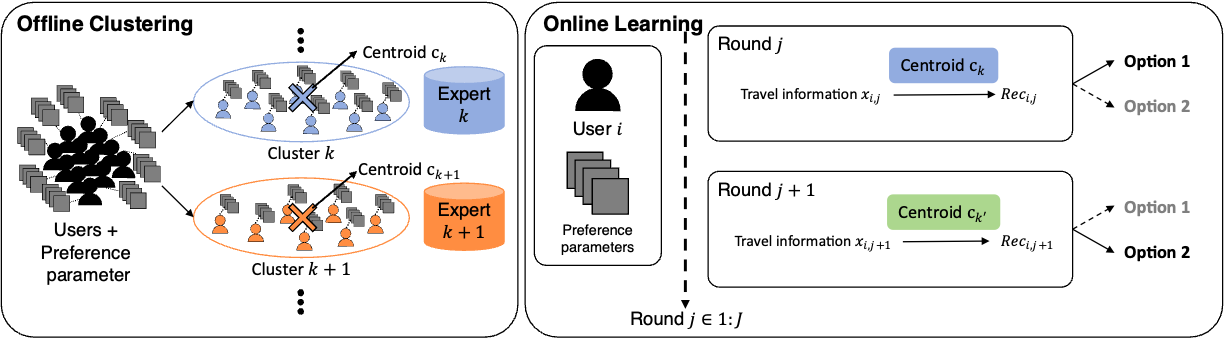}
    \caption{Illustrative figure for Expert with Clustering algorithm.}
    \label{fig:concept}
\end{figure}
\vspace{-0.1in}
\subsection{Related Works}
\textbf{Preference learning for drivers.}
The field of preference learning for drivers has focused on adaptive models that cater to individual driving behaviors and decision-making processes. \cite{jalota_online_2022} contribute to this domain with the online learning approach, adjusting traffic tolls based on aggregate flows to influence drivers towards more efficient routes, thus minimizing the need for detailed personal travel information. This method resonates with the utilitarian perspective of \cite{chorus_traveler_2009}, delving into how drivers' preference with advice is shaped by their personal preferences and perceptions of travel time uncertainty. \cite{sadigh_active_2017} explore the learning of human preferences from comparative choices between trajectories, eschewing the need for direct reward signals. \jhedit{Collectively, these works highlight a trend toward machine learning that is not just reactive but anticipatory and adaptable to the nuanced spectrum of human driving preferences.}

\noindent \textbf{Contextual bandits.}
The contextual bandits framework has emerged as an efficient approach for recommendation systems. Originally introduced by \cite{auer_firstContxtualBandits_2002}, this framework delved into the utilization of confidence bounds within the exploration-exploitation trade-off, specifically focusing on contextual bandits with linear value functions. Building upon this foundation, \cite{li_contextual-bandit_2010} expanded the application of this concept to personalized news recommendations and proposed LinUCB algorithm which has since become a benchmark in the field.

\noindent \textbf{Expert algorithm.}
The prediction with expert advice is a fundamental problem in online learning. The Hedge algorithm, also recognized as the Weighted Majority Algorithm and initially introduced by \cite{Hedge}, presents an efficient approach to addressing this challenge. 
The insights offered by the Hedge algorithm have significantly informed our development of the Expert with Clustering (EWC) framework. In terms of theoretical performance, \cite{HedgeRegretBound} has established that the Hedge algorithm's regret bound is \(O(\sqrt{T\log K})\). This regret bound provides a foundation for theoretical analysing the EWC framework.

\noindent \textbf{Contextual bandits with clustering.}
K-Means clustering, a classic unsupervised learning algorithm, was first introduced by \cite{lloyd_kmeans_1982}. With the evolution of both contextual bandits and clustering techniques, the concept of clustering users within contextual bandits was proposed by \cite{Gentile_clusterBandits_2014}, utilizing a graph to represent user similarities. To enhance the utilization of similarity information, \cite{li_collaborativeFilteringBandit_2016} combined collaborative filtering with contextual bandits to cluster both users and items. Furthermore, \cite{gentile_contextAwareClustering_2017} introduced context-aware clustering, where each item could cluster users. 
Existing works in contextual bandits typically overlook user choice due to their classical framework. In contrast, our unique problem structure incorporates user choice, offering insights into the comparative utility of different options. This distinct approach allows for accelerating the preference learning within limited data.

\noindent \revise{\textbf{Objective function in clustering.}
The classic K-Means algorithm focuses on minimizing the within-cluster sum of squares, which may not fit all clustering needs. 
For instance, K-Means assumes that clusters are spherical and roughly equal in terms of size, which may not always be the case in real-world data.
Addressing these limitations, researchers in fields such as federated learning and system identification have devised bespoke objective functions to enhance clustering methodologies.
For example, \cite{ICFA_NEURIPS2020} proposed a new framework which alternates between identifying clusters and minimizing a customized loss function for federated learning. Similarly, \cite{FMTL_IEEE2019} uses the geometric properties of loss surfaces to group clients in federated multitask learning. Additionally, \cite{clustered_system_CDC2023} applies a clustering algorithm to derive linear system models, aiming to minimize the aggregate cost function within each cluster. Building on these developments, we propose a loss-guided distance metric tailored for online preference learning.
}
\vspace{-0.1in}
\subsection{Contributions}
The primary contributions of this work are outlined as follows:
\begin{enumerate}
    \itemsep-0.5em 
    \item We introduce the novel hierarchical contextual bandit framework, Expert with Clustering (EWC), which integrates clustering and prediction with expert advice to address the online preference learning problem. This framework effectively utilizes hierarchical user information, enabling rapid and accurate learning of user preferences.
    \item \revise{We propose a distance metric, Loss-guided Distance, tailored for the online preference learning problem, which enhances the representativeness of centroids.} This advancement improves the performance of the EWC framework, demonstrating its practical effectiveness.
    \item \tyedit{We establish the regret bound of EWC as a sum of two components: the regret from the Hedge algorithm and the bias introduced by representing users with centroids,}
    indicating superior theoretical performance in the short term and enhanced overall experimental performance compared to the LinUCB algorithm (\cite{li_contextual-bandit_2010}).
\end{enumerate}
\vspace{-0.1in}
\section{Problem Formulation}
Consider the scenario where \jhedit{a social planner} is tasked with recommending \jhedit{mobility options $\mathcal{R}$, where $A := |\mathcal{R}|$, to a population of} drivers. 
Each mobility option is parameterized by a travel information vector $x_{i,t} \in \mathbb{R}^d$ \revise{specifying relevant travel metrics, where $i$ and $t$ indicates the index of driver and decision round.}
which specifies relevant travel metrics. For simplicity, we consider two mobility options ($A=2$), each with two relevant travel metrics ($d=2$), although the framework extends gracefully to more options and metrics. Thus, at each decision point for a user, \jhedit{the social planner} faces a choice between two route options: route 1, the standard route with regular travel time and emissions, and route 2, an eco-friendly alternative that, while offering reduced emissions, comes with an increased travel time. 
Intuitively, in this simplified example, the social planner seeks to quickly identify users who prefer travel time or environmental impact while ensuring user participation, in order to best achieve the system sustainability. In the future, considerations of multiple system objectives and incentives to shape user choices can be included.

For each decision round $t$, the user $i$ compare two routes in terms of their relative travel time and emissions. Let's denote the travel time and emissions for route 2 relative to route 1 as $\tau_{i,t}$ and $e_{i,t}$, respectively. For example, $[\tau_{i,t}, e_{i,t}]=[1.2, 0.9]$ means $120\%$ of travel time and $90\%$ of emission. Travel information vector for this decision round, $x_{i,t}$, is defined with two components for two routes: $x_{i,t}(1)=[1,1]$ and $x_{i,t}(2) = [\tau_{i,t}, e_{i,t}]$.
Based on this information, we issue a recommendation $Rec_{i,t}\in\{1,2\}$, whereupon we receive feedback in the form of the user's choice $y_{i,t}\in \{1,2\}$. 
The objective of our system is to minimize the total regret: $\sum_{i=1}^N \sum_{t=1}^T |Rec_{i,t}-y_{i,t} |^2$, where $N$ represent the number of users and $T$ denote the total number of decision rounds. 
\vspace{-0.1in}
\section{Expert with Clustering (EWC)}
\subsection{General Framework}
We introduce the Expert with Clustering (EWC) algorithm, a novel hierarchical contextual bandit approach. EWC transforms an online preference learning problem into an expert problem and utilizes the Hedge algorithm to identify the most effective expert.

\tyedit{Prediction with expert advice is a classic online learning problem introduced by \cite{Hedge}. Consider a scenario where a decision-maker has access to the advice of $K$ experts. At each decision round $t$, advice from these $K$ experts is available, and the decision maker selects an expert based on a probability distribution $\mathbf{p}_{t}$ and follows his advice. Subsequently, the decision maker observes the loss of each expert, denoted as $\mathbf{l}_{t} \in [0,1]^K$. The primary goal is to identify the best expert in hindsight, which essentially translates to minimizing the regret: \revise{$\sum_{t=1}^T\left(<\mathbf{p}_{t},\mathbf{l}_{t}>- \mathbf{l}_{t}(k^*)\right)$}, where $k^*$ is the best expert throughout the time.}

We cast the online preference learning problem into the framework of prediction with expert advice in the following way.
Assume that each user has a fixed but unknown preference parameter $\boldsymbol{\theta}_{i} \in \mathbb{R}^d$. Given $\boldsymbol{\theta}_{i}$, we can make predictions using a known function $\hat{y}(\boldsymbol{\theta}_{i}, x_{i,t})$. 
The EWC algorithm operates under the assumption of a cluster structure within the users' preference parameters \tyedit{$\{\boldsymbol{\theta}_{i}\}_{i\in[N]}$}.
Utilizing a set of offline training data $\mathcal{D}= \{ \{x_{i,t}\}_{i\in[N^\prime], t\in[T^\prime]}, \{y_{i,t}\}_{i\in[N^\prime],t\in[T^\prime]} \} $ where $N^\prime$ and $T^\prime$ are number of users and decision rounds in training data, we initially employ a learning framework (such as SVM or \jhedit{nonlinear regression}) to determine each user's $\boldsymbol{\theta}_{i}$. 
Despite differences between training and testing data, both are sampled from the same distribution. This allows for an approximate determination of \(\boldsymbol{\theta}_{i}\), providing insights into the hierarchical structure among users, albeit with some degree of approximation.
Subsequently, a clustering method is applied to identify centroids $\{\mathbf{c}_{k}\}_{k\in[K]}$. Each centroid is considered as an expert. Using the Hedge algorithm, we initialize their weights and, at every online decision round, select an expert $E_{i,t} \in [K]$. \revise{An expert $E_{i,t}$ provides advice suggesting that a user's preference parameters closely resemble the centroid $\mathbf{c}_{E_{i,t}}$. Consequently, we use this centroid to estimate the user’s preferences.} The recommendation \tyedit{$Rec_{i,t} = \hat{y}(\mathbf{c}_{E_{i,t}}, x_{i,t})$} is then formulated. Upon receiving the user's chosen option $y_{i,t}$, \revise{we calculate the loss for each expert and update the weights in Hedge based on this loss.} 
The loss for each expert $k$ is determined by a known loss function $\mathbf{l}_{i,t}(k) = l(\hat{y}(\mathbf{c}_{k}, x_{i,t}), y_{i,t}) \in \mathbb{R}$, e.g., $l(\hat{y}(\mathbf{c}_{k}, x_{i,t}), y_{i,t})=\mathbbm{1}_{\hat{y}(\mathbf{c}_{k}, x_{i,t})\neq y_{i,t}}$.
The details of this process are encapsulated in Algorithm \ref{Expert_with_Cluster}.
\begin{algorithm}
    \caption{Expert With Cluster}
    \label{Expert_with_Cluster}
    \begin{algorithmic}
        \REQUIRE Number of clusters $K$, offline training data $\mathcal{D}$, learning rate $\eta$
        \STATE Train with data $\mathcal{D}$, receive $\{\boldsymbol{\theta}_{i}\}_{i\in[N^\prime]}$
        \STATE Apply clustering on $\{\boldsymbol{\theta}_{i}\}_{i\in[N^\prime]}$, receive centroids $\{\mathbf{c}_{k}\}_{k\in [K]}$ 
        \STATE Initialize weight $\mathbf{p}_{i,1}(k) \leftarrow \frac{1}{K}$ for all $i\in[N], k\in[K]$
        \FOR {$t = 1,\dots,T$}
            \FOR {$i = 1,\dots,N$}
                \STATE Receive $x_{i,t}$ 
                \STATE Sample $E_{i,t} \sim \mathbf{p}_{i,t}$, submit $Rec_{i,t} = \hat{y}(\mathbf{c}_{E_{i,t}}, x_{i,t})$, 
                \STATE Receive $y_{i,t}$, compute loss $\mathbf{l}_{i,t}(k) = l(\hat{y}(\mathbf{c}_{k}, x_{i,t}), y_{i,t})$ for all $k\in[K]$
                \STATE $\mathbf{p}_{i,t+1}(k) \leftarrow \frac{\mathbf{p}_{i,t}(k)e^{-\eta \mathbf{l}_{i,t}(k)}}{\sum_{k^\prime \in [K]} \mathbf{p}_{i,t}(k^\prime)e^{-\eta \mathbf{l}_{i,t}( k^\prime)}}$ for all $k\in[K]$
            \ENDFOR
        \ENDFOR
    \end{algorithmic}
\end{algorithm}
\vspace{-0.3in}
\subsection{EWC for Online Preference Learning}
We implement the EWC algorithm for online preference learning in a driving context. \tyedit{ For each user \(i\), a linear decision boundary is posited, characterized by parameters \(\boldsymbol{\theta}_{i} = [b_i, s_i, o_i]\). Here, \(b_i\) and \(s_i\) denote the bias and slope of the line, and \(o_i\) denotes the orientation of the decision boundary that differentiates the affiliation of user's choice. $\boldsymbol{\theta}_{i}$ classifies the data points \([\tau_{i,t}, e_{i,t}]\) into two categories: opting for the regular route (\(y_{i,t} = 1\)) or the eco-friendly route (\(y_{i,t} = 2\)).}
\revise{
This example illustrates the meaning of $\boldsymbol{\theta}_i$.
Assume that the eco-friendly route is described by $x_{i,t}(2) = [\tau_{i,t}, e_{i,t}] = [1.1, 0.85]$.
Consider a user with preference parameters $\boldsymbol{\theta}_i = [b_i, s_i, o_i] = [2, -1, 1]$, which sets the decision boundary at $\tau = -e + 2$, signifying a preference for lower travel times and emissions. 
This user will opt for the eco-friendly route since it is within the decision boundary. However, if $\boldsymbol{\theta}_i = [3, -2, 1]$ which means user pays more attention to travel time, this eco-friendly route will not be preferred.
}

In the offline training phase, using the dataset \(\mathcal{D}\), a linear Support Vector Machine (SVM) is initially employed to differentiate the two classes of data points for each user \(i\). This process yields the parameters \(\{\boldsymbol{\theta}_{i}\}_{i \in [N^\prime]}\). Subsequently, K-Means clustering is applied to ascertain the centroids \(\{\mathbf{c}_{k}\}_{k \in [K]}\) of the set \(\{\boldsymbol{\theta}_{i}\}_{i \in [N^\prime]}\), where each centroid is represented as \(\mathbf{c}_{k} = [\overline{b}_k, \overline{s}_k, \overline{o}_k]\). 
\revise{$K$ serves as a hyperparameter. We select the value of $K$ that yields the minimum regret on the offline training set.}

In the online learning stage, first, the weight \(p(k)\) is initialized for each expert. For every decision instance \(t\) pertaining to user \(i\), we collect action data \(x_{i,t}\). 
Utilizing the Hedge Algorithm, an expert \(E_{i,t}\) is selected. 
The recommendation is then formulated as \(Rec_{i,t} = \hat{y}(\mathbf{c}_{E_{i,t}}, x_{i,t})\), with \(\hat{y}(\mathbf{c}_{k}, x_{i,t}) = \revise{1+}\mathbbm{1}_{\overline{o}_k (\tau_{i,t} - \overline{s}_k e_{i,t} - \overline{b}_k) > 0}\). Upon obtaining the user's choice \(y_{i,t}\), the loss \(\mathbf{l}_{i,t}(k) = |\hat{y}(\mathbf{c}_{k}, x_{i,t}) - y_{i,t}|^2\) is computed, leading to an adjustment of each expert's weight accordingly. 
\vspace{-0.1in}
\subsection{Clustering with Loss-guided Distance}
The core parameter influencing the regret in our model is the set of centroids \(\{\mathbf{c}_{k}\}_{k\in [K]}\). An accurately representative set of centroids can significantly reflect users' behaviors, whereas poorly chosen centroids may lead to suboptimal performance. In our simulations, we observed limitations with centroids generated by the standard K-Means algorithm. For instance, 
a centroid \(\mathbf{c}_{k}\) that differs slightly from \(\boldsymbol{\theta}_{i}\) in the bias term but exceeds the decision boundary can misclassify many points, resulting in higher regret. 
This implies that centroids with similar \(\boldsymbol{\theta}_{i}\) values do not necessarily yield comparable performances.
\revise{To address this issue, we introduce a distance metric guided by the loss function which is tailored for the online preference learning problem.} Our objective is to ensure that \(\theta_i\) values within the same cluster exhibit similar performance. Thus, we replace the traditional L$_2$ norm distance with the prediction loss incurred when assigning \(\mathbf{c}_{k}\) to user \(i\). Here, we define: \(\mathbf{x}_i=[x_{i,1}, x_{i,2},...,x_{i,T^\prime}] \in \mathbb{R}^{T^\prime\times A \times d}\) and \(\mathbf{y}_i=[y_{i,1}, y_{i,2},...,y_{i,T^\prime}] \in \mathbb{R}^{T^\prime}\), while \(\mathbf{\hat{y}}(\mathbf{c}_{k}, \mathbf{x}_i)= [\hat{y}(\mathbf{c}_{k}, x_{i,1}), \hat{y}(\mathbf{c}_{k}, x_{i,2}),..., \hat{y}(\mathbf{c}_{k}, x_{i,T^\prime})] \in \mathbb{R}^{T^\prime} \). The Loss-guided Distance is defined as $dist(i, \mathbf{c}_{k})=||\mathbf{\hat{y}}(\mathbf{c}_{k}, \mathbf{x}_i)-\mathbf{y}_i||^2$.
The detailed clustering is presented in Algorithm~\ref{K-Means with Loss-guided Distance}. 

\begin{algorithm}
	\caption{K-Means with Loss-guided Distance}
	\label{K-Means with Loss-guided Distance}
	\begin{algorithmic}
		\REQUIRE $\{\boldsymbol{\theta}_{i}\}_{i\in [N^\prime]}$
        \STATE Randomly initialize centroids $\{\mathbf{c}_{k}\}_{k\in[K]}$
        \WHILE{$\{\mathbf{c}_{k}\}_{k\in[K]}$ \jhedit{not converged}}
        \STATE $dist(i, \mathbf{c}_{k})\leftarrow ||\mathbf{\hat{y}}(\mathbf{c}_{k}, \mathbf{x}_i)-\mathbf{y}_i||^2$ for all $i\in[N^\prime], k\in[K]$
        \STATE $r_{i,k}\leftarrow \mathbbm{1}_{k=\arg\min_{k^\prime} dist(i,c_{k^\prime})}$ for all $i\in[N^\prime],k\in[K]$
        \STATE $\mathbf{c}_{k}\leftarrow \frac{\sum_{i=1}^{N}r_{i,k}\boldsymbol{\theta}_{i}}{\sum_{i=1}^N r_{i,k}}$ for all $k\in[K]$
        \ENDWHILE
        \RETURN $\{\mathbf{c}_{k}\}_{k\in[K]}$
	\end{algorithmic}
\end{algorithm}
\vspace{-0.3in}
\section{Regret analysis}
\subsection{Regret Bound of EWC}
\jhedit{Before describing our theoretical findings, we first introduce some background and definitions.} 
\jhedit{In the expert problem, spanning $T$ total rounds with $K$ experts, we denote the best expert throughout the duration as $k^*$. The regret bound, as established by \cite{HedgeRegretBound}, is expressed as:}
\begin{equation}
    R_{Hedge}=\sum_{t=1}^T \left( \langle \mathbf{p}_{t},\mathbf{l}_{t} \rangle -\mathbf{l}_{t}(k^*) \right) \leq 2\sqrt{T \log K}
\end{equation}
The loss of K-Means algorithm is define as $\mathcal{L}=\sum_{i=1}^N ||\mathbf{c}_{k(i)}-\boldsymbol{\theta}_{i}||^2 $, where $k(i)$ is the cluster centroid assigned to $\boldsymbol{\theta}_{i}$. \jhedit{Consider} $\{\mathbf{c}_{k}\}_{k\in[K]}$ be any set of centroids, $P$ \jhedit{as} any distribution on $\mathbbm{R}^d$ with \jhedit{mean} $\boldsymbol{\mu} = \mathbbm{E}_P[\boldsymbol{\theta}_{i}]$ \jhedit{and variance} $\sigma^2 =\mathbbm{E}_P[||\boldsymbol{\theta}_{i}-\boldsymbol{\mu}||^2]$. \jhedit{Assuming} finite Kurtosis ($4^\text{th}$ moment) $\hat{M}_4<\infty$ \jhedit{and given} $\epsilon \in (0,1)$, $\delta \in (0,1)$ and \jhedit{a sample size} $m$ from $P$, \jhedit{we establish that for} $ m\geq \frac{12800(8+\hat{M}_4)}{\epsilon^2\delta}\left(3+30K(d+4)\log 6K+\log \frac{1}{\delta}\right)$, the Uniform deviation bound of K-Means, \jhedit{as proven} by 
\cite{K-MeansBound}, \jhedit{holds} with at least $1-\delta$ \jhedit{probability:}
\begin{equation} 
    \label{K-MeansBound-Proof}
     | \mathcal{L}-\mathbbm{E}_P [\mathcal{L}]|\leq \frac{\epsilon}{2}\sigma^2 +\frac{\epsilon}{2}\mathbbm{E}_P [\mathcal{L}]
\end{equation}
\jhedit{We define} the regret of EWC as the \jhedit{performance difference} between EWC and Oracle $\boldsymbol{\theta}_{i}$:
\begin{equation}
    R_{EWC} = \sum_{i=1}^N \sum_{t=1}^T \left( \langle \mathbf{p}_{i,t}, \mathbf{l}_{i,t}\rangle - | \hat{y}(\boldsymbol{\theta}_i, x_{i,t})-y_{i,t}|^2  \right)
\end{equation}
\revise{Since the study in \cite{K-MeansBound} shows the performance of K-Means clustering using the $L_2$ norm distance, we similarly adopt the $L_2$ norm distance to analyze regret in our framework.}
What follows is our main theoretical result. 
Here we slightly abuse the notation $\mathbf{\hat{y}}(\boldsymbol{\theta}_i, \mathbf{x}_i)\in\mathbb{R}^T$ and $\mathbf{y}_{i}\in\mathbb{R}^T$ to be the prediction and user's choice vector in testing data.
\newtheorem{EWC}{Theorem}[section]
\begin{EWC}[Regret Bound of EWC]\label{EWC_Bound}
Let $P$ be any distribution of $\boldsymbol{\theta}_{i}\in\mathbbm{R}^d$ with $\boldsymbol{\mu} = \mathbbm{E}_P[\boldsymbol{\theta}_{i}]$, $\sigma^2 =\mathbbm{E}_P[||\boldsymbol{\theta}_{i}-\boldsymbol{\mu}||^2]$, and finite \textit{Kurtosis}. Let $\{\mathbf{c}_{k}\}_{k\in [K]}$ be any set of centroids, $k^*(i)$ be the  best expert for user $i$, $\mathcal{L}=\sum_{i=1}^N ||\mathbf{c}_{k^*(i)}-\boldsymbol{\theta}_{i}||^2$ be the total squared distance of clustering, and $\mathbf{\hat{y}}(\boldsymbol{\theta}_i, \mathbf{x}_i)\in \mathbb{R}^T$ be the prediction function. If $\mathbf{\hat{y}}(\cdot, \mathbf{x}_i)$ is Lipschitz continuous for all $\mathbf{x}_i$ \tyedit{with Lipschitz constant $L$, \jhedit{L$_2$} norm distance, and dimension normalization}, then with probability at least $1-\delta$, the regret of EWC is bounded by:
\begin{equation}
    R_{EWC}\leq \overline{R}_{EWC}=2N\sqrt{T\log K} +  TL\left(\frac{\epsilon}{2}\sigma^2+(\frac{\epsilon}{2}+1)\mathbb{E}_P[\mathcal{L}]\right)
\end{equation}
\end{EWC}
\vspace{-0.1in}
\begin{proof}
\begin{equation}
\begin{aligned}
    R_{EWC} =& \sum_{i=1}^N \sum_{t=1}^T \left( \langle \mathbf{p}_{i,t}, \mathbf{l}_{i,t}\rangle - | \hat{y}(\boldsymbol{\theta}_i, x_{i,t})-y_{i,t}|^2  \right)\\
    =& \sum_{i=1}^N \sum_{t=1}^T \left(\langle \mathbf{p}_{i,t}, \mathbf{l}_{i,t}\rangle - | \hat{y}(\mathbf{c}_{k^*(i)}, x_{i,t})-y_{i,t}|^2  \right) \\
    &+ \sum_{i=1}^N \sum_{t=1}^T\left( | \hat{y}(\mathbf{c_{k^*(i)}}, x_{i,t})-y_{i,t}|^2-| \hat{y}(\boldsymbol{\theta}_i, x_{i,t})-y_{i,t}|^2  \right)\\
    =& \sum_{i=1}^N \sum_{t=1}^T \left(\langle \mathbf{p}_{i,t}, \mathbf{l}_{i,t}\rangle - \mathbf{l}_t(k^*(i)) \right) + \sum_{i=1}^N \left(||\mathbf{\hat{y}}(\mathbf{c}_{k^*(i)},\mathbf{x}_i)-\mathbf{y}_{i}||^2 - ||\mathbf{\hat{y}}(\boldsymbol{\theta}_{i},\mathbf{x}_i)-\mathbf{y}_{i}||^2\right)\\
\end{aligned}
\end{equation}
\vspace{-0.1in}
\revise{By the regret bound of Hedge and triangle inequality,}
\begin{equation}
\begin{aligned}
    R_{EWC} \leq & 2N\sqrt{T\log K}+\sum_{i=1}^N||\mathbf{\hat{y}}(\mathbf{c}_{k^*(i)},\mathbf{x}_i)-\mathbf{\hat{y}}(\boldsymbol{\theta}_{i},\mathbf{x}_i)||^2\\
\end{aligned}
\end{equation}
\vspace{-0.1in}
\revise{By the Lipschitz condition,} $\exists L$ s.t. $\forall i, \forall \boldsymbol{\theta}_1,\boldsymbol{\theta}_2$, $\frac{1}{T}||\mathbf{\hat{y}}(\boldsymbol{\theta}_1, \mathbf{x}_i)-\mathbf{\hat{y}}(\boldsymbol{\theta}_2,\mathbf{x}_i)||^2\leq L ||\boldsymbol{\theta}_1-\boldsymbol{\theta}_2||^2$
    \begin{equation}
        \begin{aligned}
        \sum_{i=1}^N||\mathbf{\hat{y}}(\mathbf{c}_{k^*(i)},\mathbf{x}_i)-\mathbf{\hat{y}}(\boldsymbol{\theta}_{i},\mathbf{x}_i)||^2
            &\leq TL\sum_{i=1}^N ||\mathbf{c}_{k^*(i)}-\boldsymbol{\theta}_{i}||^2\\
            &= TL\mathcal{L}
            \leq TL (|\mathcal{L}-\mathbb{E}[\mathcal{L}]|+\mathbb{E}[\mathcal{L}])\\
            \end{aligned}
    \end{equation}
By inequation \ref{K-MeansBound-Proof}, with probability at least $1-\delta$, 
    \begin{equation}
        \begin{aligned}
        \sum_{i=1}^N||\mathbf{\hat{y}}(\mathbf{c}_{k^*(i)},\mathbf{x}_i)-\mathbf{\hat{y}}(\boldsymbol{\theta}_{i},\mathbf{x}_i)||^2
            &\leq TL\left(\frac{\epsilon}{2}\sigma^2+(\frac{\epsilon}{2}+1)\mathbb{E}[\mathcal{L}]\right)\\
        \end{aligned}
    \end{equation}
    \begin{equation}
        \begin{aligned}
        R_{EWC} &\leq 2N\sqrt{T\log K}+TL\left(\frac{\epsilon}{2}\sigma^2+(\frac{\epsilon}{2}+1)\mathbb{E}[\mathcal{L}]\right)
        \end{aligned}
    \end{equation}
\end{proof}
\vspace{-0.2in}
\revise{The Gaussian Mixture Model (GMM) aligns closely with our hypothesis of a hierarchical structure among users, which is a typical assumption in the analysis of clustering algorithms. By assuming that the distribution of users' preferences follows a GMM, we derive Corollary \ref{EWC_Bound_GMM}.}
\newtheorem{EWC_GMM}{Corollary}[EWC]
\begin{EWC_GMM}\label{EWC_Bound_GMM}
If $P$ is a Gaussian Mixture Model (GMM) with K Gaussian distributions, each of which has weight $\pi_k$, mean $\boldsymbol{\mu}_k$, and covariance $\Sigma_k$, and the clustering outputs the optimal centroids where $\mathbf{c}_{k}=\boldsymbol{\mu}_k$. Define $l_{centroids}=L\frac{\epsilon}{2N}\sigma^2+L(\frac{\epsilon}{2}+1)\sum_{k=1}^K \pi_k trace(\Sigma_k)$ be the average loss caused by centroids. With probability at least $1-\delta$, the regret of EWC is bounded by
\begin{equation}
    R_{EWC}\leq \overline{R}_{EWC}= 2N\sqrt{T\log K} +  TNl_{centroids}
\end{equation}
\end{EWC_GMM}
\begin{proof}
Since $\mathbf{c}_{k} = \boldsymbol{\mu}_k$, and $P=\sum_{k=1}^K \pi_k \mathcal{N}(\boldsymbol{\mu}_k, \Sigma_k)$, the expected squared distance is $\mathbb{E}[||\boldsymbol{\theta}_{i}-\mathbf{c}_{k(i)}||^2]=\sum_{k=1}^K \pi_k trace(\Sigma_k)$. So, \tyedit{ $\mathbb{E}[\mathcal{L}]=N\mathbb{E}[||\boldsymbol{\theta}_{i}-\mathbf{c}_{k(i)}||^2]=N\sum_{k=1}^K \pi_k trace(\Sigma_k$).}
\end{proof}
\vspace{-0.2in}
\subsection{Comparison}
We compare the regret bound of the EWC algorithm with LinUCB and oracle Follow-the-Leader (oracle FTL). \tyedit{Follow-the-Leader (FTL) is a straightforward method that selects the option with the best historical performance \revise{up to the current time step $k = \argmin_{k^\prime} \sum_{t^\prime=1}^t \mathbf{l}_{t^\prime}(k^\prime)$}.
The oracle FTL is an oracle method that lets us know \revise{the best option up to time $T$ in hindsight} and always chooses it at decision rounds.}
Lemma \ref{LinUCB_Bound} is the regret bound of SupLinUCB (a varient of LinUCB) which has been proved by \cite{li_contextual-bandit_2010}, and lemma \ref{OracleFTL_Bound} is the regret bound of oracle FTL. Corollary \ref{AdvEWC} compares EWC with both LinUCB and Oracle FTL.
\newtheorem{Adv_LinUCB}{Lemma}[subsection]
\begin{Adv_LinUCB}[Regret Bound of SupLinUCB]\label{LinUCB_Bound} 
Assume  $ \forall i,t,\exists \theta^*_i \in\mathbb{R}^d$,  s.t. $E[\mathbf{l}_{i,t}(a)|x_{i,t}(a)]=x_{i,t}(a)^\intercal \theta^*_i$. Define $R_{LinUCB}=\sum_{i=1}^N\sum_{t=1}^T \left(\mathbf{l}_{i,t}(a_{i,t})-\mathbf{l}_{i,t}(a^*_{i,t})\right)$ where $a_{i,t}^*=\arg\max_{a} x_{i,t}(a)^\intercal \theta^*$. If SupLinUCB runs with $\alpha=\sqrt{\frac{1}{2}\ln{\frac{2TK}{\delta}}}$, with probability at least $1-\delta$, $R_{LinUCB}<\overline{R}_{LinUCB}=O\left(N\sqrt{Td\ln^3{(KT\ln T/\delta})}\right)$.
\end{Adv_LinUCB}
\vspace{-0.15in}
\begin{Adv_LinUCB}[Regret Bound of Oracle FTL]\label{OracleFTL_Bound} 
Define the regret of oracle FTL be $R_{OracleFTL}=\sum_{i=1}^N\sum_{t=1}^T \mathbf{l}_{i,t}$.
Let $p_i$ be the proportion of choosing option 1 for each user $i$. The regret of oracle FTL is $R_{OracleFTL}=\sum_{i=1}^N T\min\{p_i, 1-p_i\}=O(TN)$.
\end{Adv_LinUCB}
\vspace{-0.15in}
\begin{proof}
    Since we always choose the best one of options $\{1, 2\}$ for each user, the number of wrong prediction should be $T\min\{p_i, 1-p_i\}$. So the total regret is the summation of all users.
\end{proof}
\vspace{-0.15in}
\newtheorem{Advantage of EWC}{Corollary}[subsection]
\begin{Advantage of EWC}[Advantage of EWC]\label{AdvEWC} 
1) Assume $\overline{R}_{LinUCB}= CN\sqrt{Td\ln^3{(KT\ln T/\delta})}$, then when $T< (\frac{C-2}{l_{centroids}})^2$, $\overline{R}_{EWC}<\overline{R}_{LinUCB}$. 2) When $l_{centroids}<\frac{1}{N}\sum_{i=1}^N \min\{p_i, 1-p_i\}-2\sqrt{\log (K)/T}$, $\overline{R}_{EWC}<R_{OracleFTL}$
\end{Advantage of EWC}
\vspace{-0.1in}
\begin{proof}
    1) Since $\overline{R}_{EWC}=2N\sqrt{T\log K}+TNl_{centroids}$, $\overline{R}_{EWC}<\overline{R}_{LinUCB}$ is equivalent to $\sqrt{T}l_{centroids}<c\sqrt{d\ln^3{(KT\ln T/\delta})}-2\sqrt{\log K}$. So when $\sqrt{T}l_{centroids}<c-2$, the condition above is satisfied. 2) Dividing $2N\sqrt{T\log K}+TNl_{centroids}$ and $\sum_{i=1}^N T\min\{p_i, 1-p_i\}$ by $NT$, we can get the second result.
\end{proof}
\vspace{-0.15in}
As highlighted in Corollary \ref{AdvEWC}, EWC demonstrates superior theoretical performance compared to LinUCB when $T$ is relatively small. This advantage is contingent upon $l_{centroids}$ which is the average loss incurred when using the centroids \(\mathbf{c}_{k}\) as representations of the users' preference parameters \(\boldsymbol{\theta}_{i}\). EWC outperforms the oracle Follow-the-Leader (FTL) when the loss due to employing centroids is less than the loss from consistently selecting the fixed best arm. The term \(2\sqrt{\log(K)/T}\) represents the average loss associated with the process of identifying the best expert. This loss is negligible since it decreases rapidly as $T$ increases.
\vspace{-0.15in}
\section{Experiments}
In this section, we assess the Expert with Clustering (EWC) algorithm through experiments designed to evaluate its performance in learning driver preferences online, particularly its adaptability to new data and accuracy in making eco-friendly route recommendations.
\vspace{-0.1in}
\subsection{Experimental Setup}
\noindent \textbf{Community survey.}
\jhedit{This study involved a community survey conducted in July 2023 on the University of North Carolina at Charlotte campus, and a total of 43 individuals participated.}
\jhedit{Participants provided the driving choice preferences as well as demographic data covering age, gender, ethnicity, and educational level.}
\jhedit{The survey's main component involved a series of questions assessing willingness to adhere to route recommendations under varying scenarios with distinct travel times and carbon dioxide emission levels. Participants rated their likelihood of following these recommendations in the Likert scale, offering insight into their decision-making criteria.}
\jhedit{For example, participants were asked on their likelihood to opt for an eco-friendly route offering a 10\% reduction in CO$_2$ emissions in exchange for a 5--15\% increase in travel time.}

\noindent \textbf{\jhedit{Mobility user simulation}.}
To better represent a diverse driving population, we expanded our dataset.
We use the Bayesian inference model that resembles the original distribution from the survey data (\cite{andrieu_introduction_2003}).
We refined this approach by implementing distinct utility functions for demographic segments differentiated by gender, age, and household car ownership.
Drawing samples from the posterior distribution of model parameters, we populated the dataset with individuals exhibiting a range of features and compliance behavior.
\jhedit{This methodology allowed us to produce $2000$ individual user choice records for the synthetic dataset, with parameters set at $N=800$, $N^\prime = 1200$, $T=T^\prime =40$, and $K=6$.} 
\tyedit{The optimal $K$ was selected based on regret minimization.}
The synthetic dataset features a mix of route choices that reflect various driving preferences and behaviors, providing a rich foundation for evaluating our EWC algorithm.

\noindent \textbf{Baselines.}
Our approach is benchmarked against a selection of well-established baseline algorithms.
\jhedit{\textit{Follow-the-Leader (FTL)} predicts the future actions based on historically rewarding choices. The \textit{Linear Upper Confidence Bound (LinUCB)} algorithm adapts the upper confidence bound method for linear payoffs, optimizing the trade-off between exploring new actions and exploiting known ones. \textit{Oracle Follow-the-Leader (Oracle FTL)} always chooses the historically optimal option. The \textit{Oracle Cluster} algorithm makes predictions using precise cluster assignments to incorporate collective behaviors within a user's group for decision-making. Lastly, \textit{Oracle $\boldsymbol{\theta}_{i}$} leverages a perfect understanding of user preferences and behaviors to anticipate the most likely user action. 
}
\vspace{-0.1in}
\subsection{Results}
\begin{figure}[!t]
    \centering    
    \includegraphics[width=0.6\textwidth]{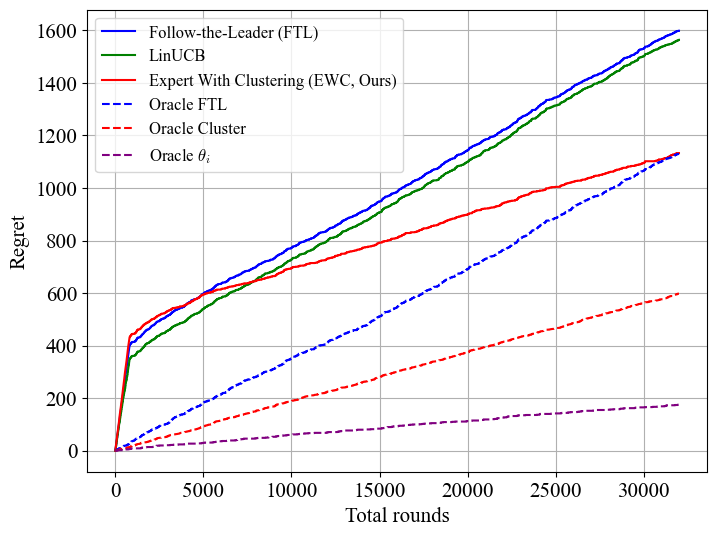}
    \caption{Comparative regret analysis of online learning algorithms: Expert with Clustering (EWC, Ours) shows lower regret than the baseline algorithms (Follow-the-Leader, LinUCB, Oracle FTL) and approaches the consistency of the Oracle methods.}
    \label{fig:temp-result}
\end{figure}

Figure~\ref{fig:temp-result} compares the regret of various online learning algorithms on a synthesized dataset based on driving preferences and eco-driving recommendations.
Regret measures how much worse an algorithm performs compared to the best possible action over time.
Oracle $\theta_i$ shows the lowest regret, indicating it almost perfectly predicts the best action due to its assumption of perfect user preference information. 
Oracle Cluster also performs well, benefiting from knowledge of user clusters. 
Oracle FTL exhibits a high slope value comparable to that of standard FTL.
LinUCB and FTL algorithms experience higher regret.
\tyedit{FTL, relying solely on historical action frequency, performs worst among baselines. LinUCB's expressiveness is limited, which leads to a similar performance with FTL.}
In the early rounds, LinUCB and the EWC algorithm start with similar levels of regret, suggesting that initially, both algorithms perform comparably in predicting the best action. 
This could be because, in the initial stages, there's less historical data to differentiate the predictive power of the algorithms, or the correct action is more obvious.
\tyedit{EWC surpasses non-oracle methods, showing clustering's effectiveness in capturing user preferences. Its long-term regret slope mirrors that of Oracle Cluster, suggesting rapid identification of optimal user group affiliations.}
\jhedit{The EWC algorithm's performance gradually improves, indicating that it is increasingly predicting the optimal actions, reducing regret by 27.57\% compared to the LinUCB at the final rounds.
This could be due to the inherent advantages of its clustering-based predictive model, which, despite lacking perfect foresight, benefits from insights that approach the prescience of the Oracle methods.}
\vspace{-0.2in}
\section{Conclusion}
In this paper, we introduce Expert with Clustering (EWC), a novel hierarchical contextual bandits algorithm designed to address the online learning challenges for drivers' preferences. EWC uniquely combines clustering techniques with prediction based on expert advice, effectively achieving low regret in online learning scenarios. 
Furthermore, EWC offers an efficient method for extracting insights into both population-wide and individual-specific behaviors, proving particularly effective in contextualized settings that exhibit hierarchical structures.
In future work, we plan to refine EWC by incorporating more user-specific preference learning and investigating the preference for incentives, thereby enhancing the personalization and effectiveness of our recommendations.
\vspace{-0.1in}
\acks{This work was partially supported by the National Science Foundation (NSF) under grant number 2149511 and the Kwanjeong scholarship.
The authors would like to thank Prof. Christos Cassandras, Prof. Andreas Malikopoulos, and Prof. Roy Dong for the insightful discussion.}
\appendix
\bibliography{egbib}

\end{document}